\documentclass[journal]{IEEEtran}

\usepackage{times}
\usepackage{epsfig}
\usepackage{graphicx}
\usepackage{amsmath}
\usepackage{amssymb}
\usepackage{tabularx, rotating}
\usepackage{float}
\usepackage{hyperref}
\usepackage{parskip}

\usepackage{array}
\usepackage{multirow}
\usepackage[utf8]{inputenc}
\usepackage{xcolor}
\usepackage[ruled,vlined]{algorithm2e}

\begin{document}

\title{Transforming Facial Weight of Real Images \\by Editing Latent Space of StyleGAN}

\author{V~N~S~Rama~Krishna~Pinnimty\textsuperscript{\textsection}, Matt~Zhao\textsuperscript{\textsection}, Palakorn~Achananuparp, and Ee-Peng~Lim

\thanks{
V.N.S.R.K. Pinnimty, M. Zhao, P. Achananuparp, and E.-P. Lim are with the School of Information Systems, Singapore Management University. Email: \{ramap, mattzhao, palakorna, eplim\}@smu.edu.sg}
}

\maketitle
\begingroup\renewcommand\thefootnote{\textsection}
\footnotetext{Equal contribution}
\endgroup

\section*{abstract}
\label{sec:abstract}
We present an invert-and-edit framework to automatically transform facial weight of an input face image to look thinner or heavier by leveraging semantic facial attributes encoded in the latent space of Generative Adversarial Networks (GANs). Using a pre-trained StyleGAN as the underlying generator, we first employ an optimization-based embedding method to invert the input image into the StyleGAN latent space. Then, we identify the facial-weight attribute direction in the latent space via supervised learning and edit the inverted latent code by moving it positively or negatively along the extracted feature axis. Our framework is empirically shown to produce high-quality and realistic facial-weight transformations without requiring training GANs with a large amount of labeled face images from scratch. Ultimately, our framework can be utilized as part of an intervention to motivate individuals to make healthier food choices by visualizing the future impacts of their behavior on appearance.

\section{Introduction}
\label{sec:introduction}
%

People tend to be less motivated to adopt healthy lifestyles, especially when the consequences of their behaviors are long-term and inconspicuous. For example, young adults who excessively consume diets which are high in calories and added sugar may not develop type-2 diabetes until decades later. Interventions designed to provide information are shown to be ineffective in motivating meaningful behavior change \cite{Whitehead_2012_AJPH,Achananuparp_2018}. On the other hand, Appearance-based behavioral interventions, involving emphasizing the impacts of behavior on appearance, have shown to be more effective in motivating behavior changes than traditional information-provision based interventions \cite{Whitehead_2012_AJPH}. Among various forms of appearance-based signals, \textit{facial weight} (also called \textit{facial adiposity}) has proven to be a good predictor of health and health outcomes, including obesity and type-2 diabetes \cite{Henderson2016}. Thus, an automated tool for simulating changes in facial weight of individuals based on their dietary behavior of future food choices may prove to be a useful feature in an appearance-based dietary intervention.

Existing approaches to transforming the facial weight of face images typically employ computer graphics and image processing techniques, such as 2D face morphing  \cite{Henderson2016} and 3D face reconstruction and face reshaping \cite{zhao_jin_2018,Xiao_2020_MM}. Most of these techniques only work well on face images in highly constrained conditions (e.g., frontal pose, neutral expression, and plain background) \cite{Henderson2016} and often require manual efforts \cite{Henderson2016,zhao_jin_2018} to generate optimal results, making them impractical for societal-scale interventions. 

To overcome the problems, we propose a framework based on a recent invert-and-edit approach \cite{guan_2018,Shen_2020_CVPR} to incrementally transform the facial weight of a real image by manipulating the image in GAN latent space. By leveraging the latent space of the state-of-the-art StyleGAN \cite{Karras_2019_CVPR} which encodes rich semantics of human faces, the proposed framework is able to effectively handle face images with diverse characteristics in both constrained and unconstrained conditions, producing visually compelling transformations without additional manual efforts.
Our work contributes to existing research by: (1) exploring the task of progressive/regressive transformation of facial weight by real image editing in latent space; (2) empirically demonstrating the feasibility and practicality of the proposed framework on diverse sets of face images.

\section{Related Work}
\label{sec:related_work}
In general, real image manipulations with GANs can be categorized into two major approaches: \textit{image-to-image translation} \cite{Isola_2017_CVPR,Zhu_2017_ICCV,Choi2018} and \textit{image editing in latent space} \cite{Jahanian_2020_ICLR,Shen_2020_CVPR}. In the former, the goal is to learn the mapping from an input domain to an output domain \cite{Isola_2017_CVPR,Zhu_2017_ICCV} or all mappings among multiple domains \cite{Choi2018}. Whereas, the latter exploits linear interpolations between encoded images in GAN latent space \cite{DCGAN_radford2015,Karras_2019_CVPR} by uncovering and navigating along latent-space directions which correspond to changes in visual attributes. Supervised \cite{guan_2018,Shen_2020_CVPR} and self-supervised \cite{Jahanian_2020_ICLR} approaches have been explored to find latent-space directions for image editing.

Our work extends beyond existing face image manipulation work \cite{Chen_2016_NIPS,Lample_NIPS2017,Choi2018,guan_2018,Xiao_2018_ECCV,Karras_2019_CVPR,Shen_2020_CVPR} that typically focuses on facial attributes, such as pose, gender, age, expression, bangs, hair color, mouth opening, and eyeglasses. To our knowledge, facial weight progression and regression via latent-space editing has not been explored before. Our framework is built on an \textit{invert-and-edit} approach \cite{richardson2020encoding}, such as InterFaceGAN \cite{Shen_2020_CVPR}, which involves obtaining inverted latent codes of input images \cite{Abdal_2019_ICCV,Zhu_2020_ECCV} and uncovering attribute directions in latent space of pre-trained GANs via supervised learning  \cite{guan_2018,Shen_2020_CVPR}. We further contribute to prior research by empirically evaluating various qualities of the generated images. Lastly, our work and \cite{Xiao_2020_MM} share similar goals, however, their approach is mostly based on 3D face reconstruction and reshaping. Compared to theirs, ours is more extensible, allowing for multi-attribute transformations without retraining.

\section{Methodology}
\label{sec:methods}
Our proposed framework leverages the state-of-the-art pre-trained image generation model StyleGAN \cite{Karras_2019_CVPR} to create highly realistic facial-weight transformations. Specifically, our goal is to use StyleGAN generator to produce an incremental change in facial weight of an arbitrary input face image from its manipulated latent code.

The framework consists of three main steps. First, the input image is pre-processed to extract and align the face region (see Section~\ref{sec:pre_processing}). This results in an aligned face image with $1024\times1024$ resolution. Next, the aligned face image is embedded into the StyleGAN manifold to produce a corresponding latent representation (see Section~\ref{sec:latent_embedding}). We use StyleGAN's extended latent space $W^+$ for embedding and obtain the inverted latent code as a concatenation of $18$ different $512$-dimensional  \textit{w} vectors. Lastly, we extract a $18\times512$ dimensional facial-weight attribute vector via supervised learning, algebraically combine the latent code with the extracted attribute vector, and pass the edited latent code to StyleGAN generator to obtain the transformed image (see Section~\ref{sec:weight_transformation}).

\subsection{Pre-processing}
\label{sec:pre_processing}

In real-world applications, users may supply input images that are not readily suitable for processing by our pipeline.
The input image may be of arbitrary resolution, may contain multiple faces, or may not even have any faces. To tackle such issues, we design a robust pre-processing pipeline consisting of the following sequential steps:

\textbf{Face Detection.}
We use Max-Margin Object Detection (MMOD) model in the Dlib Python package, which is effective in detecting faces from images even for those with some degree of rotation.
When multiple faces are detected, we only keep the primary subject's face (the largest bounding box). The input image will then be cropped around the bounding box of the detected face.

\textbf{Facial Landmarks Extraction.} We use a well-known $68$-facial landmark model implemented the Dlib Python package to detect landmarks. Additionally, we use these landmarks to calculate auxiliary parameters like eye-to-eye distance, centroid of the eyes, etc., and compute the angle required for applying face de-rotation.

\textbf{Face De-rotation.}
To ensure a high-quality transformation from our pipeline, the input face image is required to be at a near zero degree in-plane rotation. For this, we \textit{warp} and \textit{transform} the input face image to a coordinate space where: faces are centered, eyes lie on a horizontal line, and size of all the faces are approximately identical. Furthermore, we construct a transformation matrix and apply affine transformation to de-rotate the image.

\textbf{Image Alignment.} Lastly, to adjust a given face image into StyleGAN's canonical face position, we apply the same data preparation steps used in Karras et al. \cite{Karras_2019_CVPR} for padding, shrinking, or up-scaling. After this step, we get an image with $1024\times1024$ resolution that is used for latent space embedding.

\subsection{Latent Space Embedding}
\label{sec:latent_embedding}

We adapt the state-of-the-art optimization-based embedding method Image2StyleGAN \cite{Abdal_2019_ICCV} to invert real images to StyleGAN latent codes. In particular, we further modify the initialization step and the loss functions to improve embedding quality and run-time efficiency. Our embedding algorithm, based on StyleGAN-Encoder \cite{baylies_2019}, is shown in Algorithm~\ref{algo:embedding}.

\newcommand\mycommfont[1]{\footnotesize\ttfamily\textcolor{blue}{#1}}
\SetCommentSty{mycommfont}
\SetKwInput{KwInput}{Input}
\SetKwInput{KwOutput}{Output}

\begin{algorithm}[ht!]
\caption{Improved Latent Space Embedding}
\label{algo:embedding}

\DontPrintSemicolon
\KwInput{Pre-processed image $I$ $\in$ $\mathbb{R}^{1024 \times 1024 \times 3}$; \\ \hspace{10mm} gradient descent update $F^\prime(.)$; pre-trained\\ \hspace{10mm} $ResNet50$ model; learning rate $\eta$}
\KwOutput{Optimal latent code $w^{*}$; embedded\\ \hspace{12.5mm} image $G(w$) optimized via $F^\prime$}
$w$ $\leftarrow$ $ResNet50$($I$)\\
$loss_{min}$ = $\infty$\\
\While{iteration = $1$ \dots E}{
    $L$ $\leftarrow$ $L_{vgg}(G(w)$, $I$) + $L_{mse}(G(w)$, $I$)\\
    $w$ $\leftarrow$ $w$ - $\eta$ $F^\prime(\nabla_{w}.L)$\\
    \If{$L$ $<$ $loss_{min}$}{
        $w^{*}$ = $w$\\
        $loss_{min}$ = $L$
    }
}

\end{algorithm}

\textbf{Initialization.}
First, we start by feeding the aligned face image \textit{I} to a pre-trained $ResNet50$ \cite{He2016} model to extract the initial latent code. This latent code, when passed through the StyleGAN generator, gives a corresponding embedded image \textit{G($w$)}. We chose $ResNet50$ for its ability in learning better low-level features and its faster rate of convergence over $VGG16$ and $VGG19$.
Compared to random initialization or mean-face initialization \cite{Abdal_2019_ICCV} strategies, this strategy tends to produce inverted latent codes with higher reconstruction quality (see supplementary material) and can quickly converge within the specified number of epochs, achieving a good trade-off between quality and run-time efficiency.

\textbf{Optimization.}
Starting from an initial latent code $w$, we aim to arrive at the closest possible approximation of the input image in the latent space. We perform gradient descent-based optimization using a weighted loss function over a fixed number of iterations. 
Inspired by \cite{Abdal_2019_ICCV}, we use the same VGG and pixel-wise MSE loss combination with the only difference in the choice of the layers used for calculating the VGG loss. Our proposed loss function is as follows:

\begin{equation}
\resizebox{0.9\hsize}{!}{
$w^{*} = arg\hspace{1mm}min_w\hspace{1mm}
\lambda_{vgg} \cdot L_{vgg}(G(w), I)\hspace{1mm} + 
\hspace{1mm}\lambda_{mse} \cdot L_{mse}(G(w),I)$
}
\end{equation}

where $w^*$ is the optimal latent code; $\lambda_{vgg}$ is the scalar used to assign weight to the VGG perceptual loss; $G(.)$ is the pre-trained StyleGAN generator; $w$ is the latent code to optimize; $I \in \mathbb{R}^{n\times n\times 3}$ is the input image; $\lambda_{mse}$ is the scalar used to assign weight to the pixel-wise MSE loss.

To get optimal results, we set $\lambda_{vgg}=1$, $\lambda_{mse}=1$,  $n=256$ (i.e., $I \in \mathbb{R}^{256\times 256\times 3}$) when calculating VGG perceptual loss, and $n=1024$ (i.e., $I \in \mathbb{R}^{1024\times 1024\times 3}$) when calculating pixel-wise MSE loss. 

For VGG loss, we use a single-layer loss involving the $conv3\_2$ layer (layer-$9$ of VGG$16$) instead of the multi-layer loss involving $conv1\_1$, $conv1\_2$, $conv3\_2$, and $conv4\_2$ VGG$16$ layers in the original Image2StyleGAN. Visually, we observed that multi-layer VGG loss did not significantly affect the overall quality of the embedded face images. 
Formally, our VGG perceptual loss ${L_{vgg}}$ is defined as follow:

\begin{equation}
    L_{vgg}(G(w), I) = \frac{1}{N_{9}}||F_{9}(G(w))-F_{9}(I)||^{2}_{2}
\end{equation}

where $N_{9}$ is the number of scalars in the output of $conv3\_2$ layer of VGG$16$; $F_{9}$ is the feature output of $conv3\_2$ layer of VGG$16$.

We use L$2$-norm for measuring the difference between the pixels. Thus the pixel-wise MSE loss is defined as:
\begin{equation}
L_{mse}(G(w),I)=\frac{1}{N}||G(w)-I||^{2}_{2}
\end{equation}

where $N$ is the number of scalars in the image (i.e., $N = n \times n \times 3$).

\subsection{Facial-Weight Transformation}
\label{sec:weight_transformation}

The final step in our framework is to manipulate the optimal latent code $w^*$ so that it can be fed into the StyleGAN generator to produce the desired facial-weight transformation. To achieve that, we adopt a general approach similarly employed in \cite{guan_2018, Shen_2020_CVPR}, which consists of the following steps:

\textbf{Features Extraction.}
First, we aim to uncover a hyperplane in the StyleGAN latent space that separates samples into two facial-weight categories, i.e., thin and heavy. This is achieved by training a supervised facial-weight attribute classifier.

We constructed a thin/heavy labeled images dataset by generating 10K synthetic face images along with their latent codes using StyleGAN. After discarding images with noisy artifacts and irregularities, 9.9K images (StyleGAN-9.9K) were kept. Next, each image was manually assigned either a thin (4K) or a heavy (5.9K) class label by one of the co-authors of this paper. Using the manually labeled dataset, we trained a logistic regression classifier to predict a thin/heavy label $\hat{y}$ from a $18\times512$ dimensional latent code $w^*$.

\begin{equation}
    \hat{y}=f(w^*)=\frac{1}{1+e^{-(a \cdot w^* + b)}}
\end{equation}

where a vector parameter $a$ is the desired \textit{facial-weight attribute vector} representing the attribute \textit{direction} in $w^*$.

As StyleGAN latent space is not perfectly disentangled, manipulating $w^*$ along the facial-weight direction $a$ may inadvertently affect other correlated attributes. To better control the transformation \cite{Shen_2020_CVPR,guan_2018}, we perform \textit{projection subtraction} to find a projected facial-weight attribute vector $a-proj_{x}a$ where $x$ is an attribute direction to be disentangled from $a$. Given $n$ correlated directions; $X=\{x_1, x_2, ..., x_n\}$, we repeat the projection subtraction one direction at a time.

\textbf{Latent Space Manipulation.}
Given the projected facial-weight attribute vector, we manipulate the facial-weight attribute of the latent code $w^*$ as follow:

\begin{equation}
w^*_{edit}=w^* + \alpha \cdot a
\end{equation}
where $w^*_{edit}$ is the edited latent code which when passed through the StyleGAN generator produces transformed images, $w^*$ is the optimal latent code, $\alpha$ is the scalar used to control the degree of transformation towards thinner ($\alpha<0$) or heavier ($\alpha>0$) faces, and $a$ is the $18\times512$ dimensional projected facial-weight attribute vector. We only apply the editing operation to the first 8 layers of $w^*$ as we found them to be the most pertinent layers to facial weight.

\section{Experiments}
\label{sec:experiments}
To measure the performance of our framework, we present experimental evaluations on two sets of face images with varied visual attributes. First, we quantitatively measure: (i) the reconstruction quality of the latent space embedding; and (ii) the visual quality and identity-preserving quality of the transformations. Then, we visually examine examples of transformed images in a qualitative evaluation. Lastly, we assess the realism of the transformations through human evaluation.


\subsection{Experimental Setup}
\label{sec:exp_setup}

\textbf{Datasets.}
We manually selected high-resolution face images of real people from two existing datasets: 100 images from Chicago Face Database (CFD-100) \cite{Ma2015} and 100 images from WIDER FACE test set (WIDER-100) \cite{yang2016wider}, as our test datasets. 

The original CFD dataset contains images from 597 subjects of Asian, Black, Latino, and White ethnic backgrounds. All CFD images were taken in a constrained condition, i.e., straight frontal pose, neutral facial expression, and plain background. Our CFD-100 samples comprise 30 Asian, 20 Black, 30 Latino, and 20 White subjects with a balanced gender distribution across all groups.

In contrast, the WIDER FACE test images were taken in unconstrained conditions (i.e., ``in the wild") with a wide variety of scales, poses, occlusions, expressions, makeups, and illuminations. Additionally, the 100 selected images were manually annotated by one of the co-authors to identify attributes such as gender, age group, ethnicity, facial expression, and angle. The samples consist of subjects with near-uniform splits of genders (50 female and 50 male), age groups (45 young and 55 middle-age or older), ethnicity (55 White and 46 non-White), expressions (34 neutral and 66 non-neutral), and angles (56 frontal and 44 non-frontal). We expect CFD-100 to produce better overall results than WIDER-100.

\textbf{Implementation Details.} 
For all our experiments, we used StyleGAN trained on $1024\times1024$ resolution Flickr-Faces-HQ images (StyleGAN-FFHQ) \cite{Karras_2019_CVPR}. 

In the latent space embedding step, we used a pre-trained $ResNet50$ encoder, trained on a dataset of 20k StyleGAN generated face images \cite{baylies_2019},
to obtain the initial latent code. Next, we used Adam optimizer with the following optimal hyperparameters: learning rate $\eta = 0.01$, $\beta_1 = 0.9$, $\beta_2 = 0.99$, and $\epsilon = 1e^{-8}$. Moreover, we set the number of iterations $E=1000$ (Algorithm~\ref{algo:embedding}).
On average, it took approximately $1.25$ minutes to invert one image on a $32$GB Nvidia Tesla V$100$ GPU, compared to 7 minutes when using \cite{Abdal_2019_ICCV}.

Prior to the facial-weight transformation step, we examined potential entanglements between the facial-weight attribute and other facial attributes. Using 200K labeled face images from CelebA dataset \cite{liu2015faceattributes} with \textit{age}, \textit{gender}, and \textit{mouth-opening expression} attributes, we trained a binary classifier for each attribute. Given the attribute classifiers, we followed similar procedures in Section~\ref{sec:weight_transformation} to uncover the corresponding attribute directions in StyleGAN latent space and measured their correlations with the facial-weight direction using cosine similarity (see supplementary material for details). Next, we performed projection subtraction to disentangle the facial-weight attribute direction from the mouth-opening expression direction (i.e., the most correlated attribute) and used the projected direction for editing.

\subsection{Quantitative Evaluation} 
\label{sec:exp-quant}

\textbf{Evaluation Metrics.} 
Firstly, to measure the reconstruction quality of the embedded images, we use two standard perceptual metrics: \textit{peak signal-to-noise ratio} (PSNR $\in [0, \infty)$), \textit{structural similarity} (SSIM $\in [0, 1]$), and \textit{perceptual similarity metric} with AlexNet (LPIPS $\in [0, 1]$) \cite{zhang2018perceptual}. Higher PSNR and SSIM scores suggest better reconstruction quality, whereas higher LPIPS scores indicate lower reconstruction quality. Among these metrics, LPIPS is most consistent with human perception \cite{zhang2018perceptual}. For each test image, we obtained an aligned image output after the pre-processing step and computed the scores for all 200 aligned-embedded image pairs.

Next, we measure the perceptual quality and the identity preservation aspects of the transformations using \textit{Fréchet Inception Distance} (FID $\in [0, 1]$) and \textit{Openface face recognition} scores (FR $\in [0, 4]$) \cite{amos2016openface}, respectively. In general, lower FID scores indicate higher visual quality. Similarly, lower FR scores suggest that the subject's original identity is more preserved after the transformation. For each dataset, we first generated 5K transformed images from 200 test images using 50 different $\alpha$ values in [-5, 5] range. Then, we computed the FID between a reference set of 200 embedded images and the generated set of 5K images. For FR, we generated four thinner/heavier transformed images using $\alpha=\{-5, -3, 3, 5\}$ for each test image and calculated the scores for 800 aligned-transformed image pairs from both datasets.

\textbf{Results.}
We first examine the reconstruction quality by measuring the similarity between the input real images and the embedded images. As shown in Table~\ref{tbl:scores}, our framework produced better quality embedded images for CFD-100 dataset than WIDER-100 dataset according to all metrics. The results are as expected since CFD-100 data are more visually standardized and less noisy than WIDER-100 data.

\begin{table}[th]
\centering
\caption{Quantitative Evaluation Results}
\label{tbl:scores}
\resizebox{\linewidth}{!}{
\begin{tabular}{lrrlll}
\hline
 & \multicolumn{3}{c}{Latent Space Embedding} & \multicolumn{2}{c}{Transformation} \\ \cline{2-6} 
 & \multicolumn{1}{l}{PSNR (dB) ($\uparrow$)} & \multicolumn{1}{l}{SSIM ($\uparrow$)} & LPIPS ($\downarrow$) & FID ($\downarrow$) & FR ($\downarrow$) \\ \hline
CFD-100 & 32.988 & 0.764 & 0.213 & 15.392 & 0.218 \\
WIDER-100 & 31.625 & 0.747 & 0.312 & 33.98 & 0.392 \\ \hline
\end{tabular}
}
\end{table}

Next, the mean FID scores in Table~\ref{tbl:scores} indicate that the transformed CFD-100 images have higher perceptual quality than those of WIDER-100. For identity preservation, we first excluded cases in which aligned images failed to be reconstructed properly by manually checking candidate embedded images with FR $\geq$ 1. As a result, we removed 8 poorly embedded images and 40 corresponding transformed images from WIDER-100. No such failure cases were found in CFD-100 images. After data filtering, the mean FR scores in Table~\ref{tbl:scores} suggest that the identity of CFD-100 images are more preserved after the transformations than those of WIDER-100. Overall, the results are consistent with our expectation. 

\subsection{Qualitative Evaluation}
\label{sec:exp-qual}

\begin{figure*}[th]
\centering
\includegraphics[width=.7\textwidth]{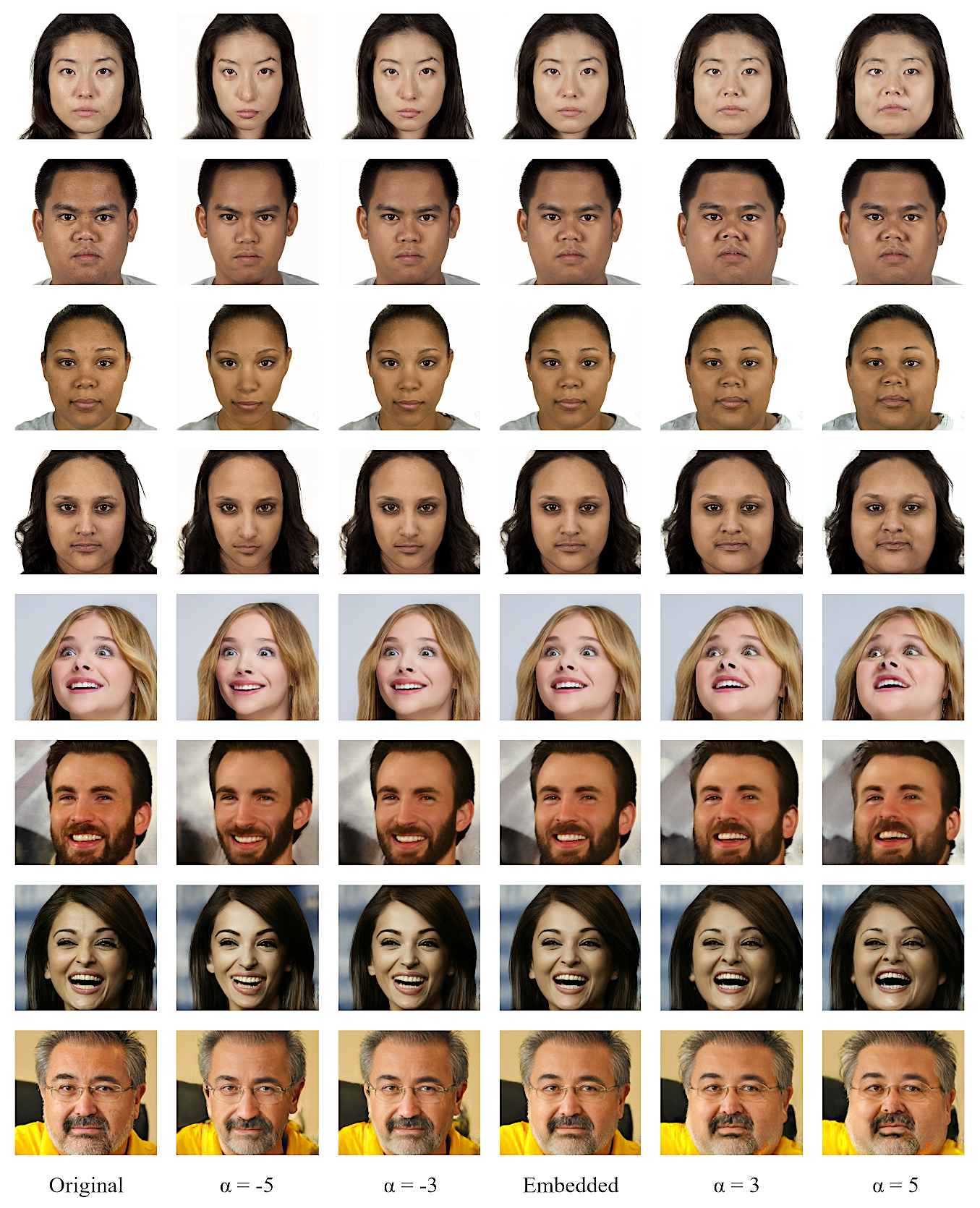} 
\caption{\textit{Facial-Weight Transformation Results.} Columns 1 and 4 show the original and the embedded images, respectively. Columns 2-3 and 5-6 display the thinner ($\alpha=-5$ and $\alpha=-3$) and heavier ($\alpha=3$ and $\alpha=5$) transformations, respectively. Rows 1-4 and 5-8 correspond to samples taken from CFD-100 and WIDER-100 datasets, respectively. Zoom in for better resolution.}
\label{fig:transformation_results}
\end{figure*}

\textbf{Facial-Weight Transformations.}
Fig.~\ref{fig:transformation_results} displays eight selected samples of original 1024$\times$1024 resolution input images (column 1) from CFD-100 (rows 1-4) and WIDER-100 (rows 5-8) datasets and their corresponding embedded (column 4) and transformed images (columns 2-3 and 5-6). As we can see, our framework generates high quality and realistic results. Not only does it produce progressive/regressive changes in specific features (i.e., cheeks, chin, and neck) and facial characteristics (i.e., mouth curvature) during weight gain/loss \cite{Henderson2016}, but it also preserves their identity and ethnicity. These realistic transformations were enabled by the encoded information in StyleGAN latent space without the need for explicit face reshaping functions \cite{Xiao_2020_MM} (see supplementary material for comparisons). Moreover, the latent facial-weight direction is shown to be independent of the natural face shapes. For instance, subjects with a square face shape (rows 6 and 8) were transformed to look heavier without having their faces becoming completely round. Additionally, it is able to generalize for different face angles, e.g., frontal (rows 1-4), 3/4 view (rows 5-7), and upward tilt (row 5). Lastly, a variety of facial expressions are well preserved during the transformation, e.g., neutral (rows 1-4), smiling (rows 5-6), and laughing (row 7). More examples can be found in the supplementary material.

\begin{figure}[ht]
\centering
\includegraphics[width=0.6\linewidth]{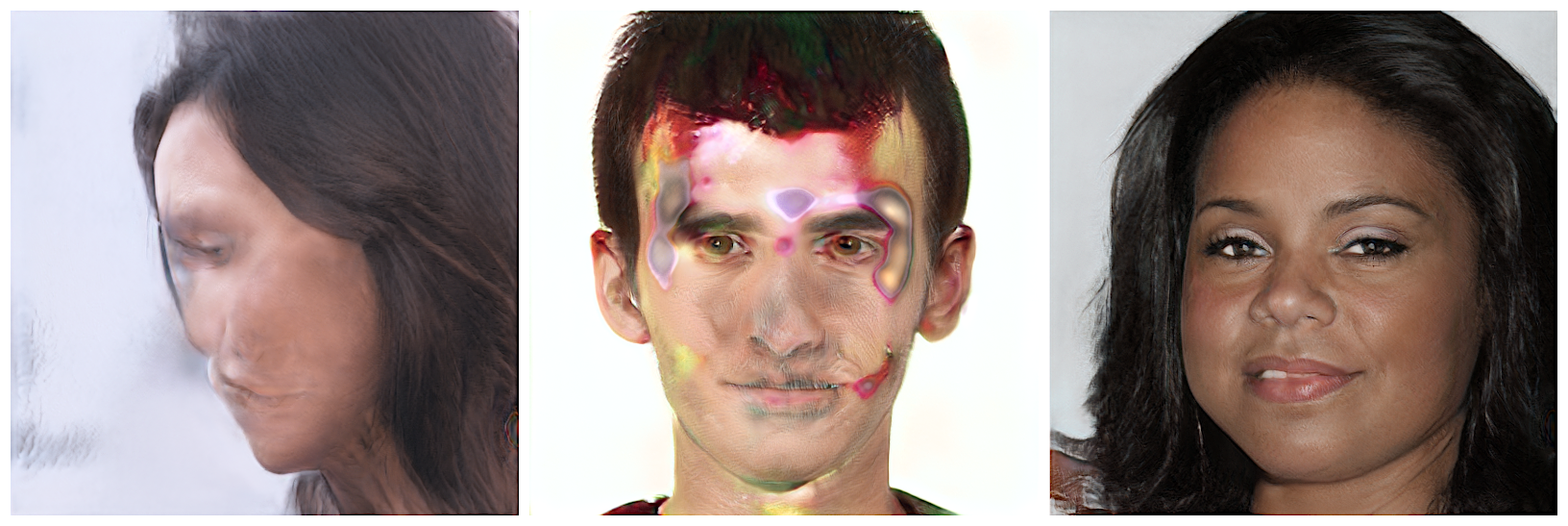} 
\caption{Examples of common failure cases such as facial deformity, blob artifacts, and entangled features, respectively.}
\label{fig:failures}
\end{figure}

\textbf{Failure Cases.}
Fig.~\ref{fig:failures} shows three examples of common failure cases that highlight the limitations of our framework. The first image depicting a subject with a thinner-transformed  ($\alpha=-3$) side-profile face pose suggests the StyleGAN-FFHQ's limit in generalizing beyond frontal and 3/4-view face poses, resulting in facial deformity. One way to handle this out-of-distribution shapes issue is to augment the training data of StyleGAN-FFHQ with more diverse face angles. Next, blob-like artifacts tend to appear in approximately 5\% of all generated faces, especially those with a large $\alpha$ value (e.g., $\alpha=-5$ in the second image). This inherent problem has been improved in StyleGAN2 \cite{Karras_2020_CVPR}. The last image shows that, in some small cases, facial weight is still entangled with a mouth-opening expression even though projected attribute vector was already used in editing. Lastly, we observed noticeable artifacts and feature distortions (e.g., elongated or tilted-up noses) in some images, e.g., row 1 in Fig.~\ref{fig:transformation_results}.

\subsection{Human Evaluation}
\label{sec:exp-amt}

\textbf{Setup.} 
We conducted a crowdsourced user study on Amazon Mechanical Turk (AMT) to investigate how humans perceive the realism of our facial-weight transformations. First, we generated thinner ($\alpha=\{-3,-5\}$) and heavier transformations ($\alpha=\{3,5\}$) for a set of 200 face images (100 StyleGAN-generated and 100 real face images). Given this set of images, we submitted 200 corresponding human intelligence tasks (HITs) to AMT. Each task requires AMT workers to sort a randomly shuffled sequence of five images, i.e., the subject's original image and four of his/her generated images by facial weight from the thinnest to the heaviest (similar to those in Fig.~\ref{fig:transformation_results}). Each task was assigned to three AMT workers, resulting in 600 responses. We further discarded 30 responses due to data input errors made by some workers.

\textbf{Results.}
According to 570 crowdsourced responses, our facial-weight transformed images are highly realistic. A vast majority of responses (71.4\%) gave the exact ordering of image sequences. In addition, 87.8\% of responses correctly identified the thinner transformed images as having lower facial weights than the original subjects. Likewise, 85.2\% of responses found the heavier transformed images to be of higher facial weights than the original images. Lastly, small percentages of incorrect responses (13.6\% - 15.3\%) show the difficulty in distinguishing similar-weight faces, e.g., $\alpha=3$ vs. $\alpha=5$.

\section{Conclusion}
\label{sec:conclusion}
Motivated by appearance-based health intervention, we propose a framework for transforming facial weight of real images by inverting and editing the input images in StyleGAN latent space. 
Next, we conducted comprehensive experiments to evaluate the performance of our framework using two face images datasets comprising subjects from a diverse demographic backgrounds and visual attributes. The results suggest that not only is our framework capable of producing facial-weight transformed images with high visual quality and realism, it is also effective in preserving the identity and characteristics of subjects after the transformations.


\textbf{Acknowledgments.} This research is supported by the National Research Foundation, Singapore under its International Research Centres in Singapore Funding Initiative. Any opinions, findings and conclusions or recommendations expressed in this material are those of the author(s) and do not reflect the views of National Research Foundation, Singapore.

{\small
\bibliographystyle{IEEEtran}
\bibliography{main.bib}
}

\section*{Appendix}
In this section, we present additional analyses and results supplementary to the main paper, including: 
\begin{itemize}
    \item Quantitative and qualitative analyses of the \textit{ResNet50} and mean-face initialization strategies used at the start of the latent space embedding step
    \item Analysis of potential entanglement between the feature-weight attribute and others
    \item Additional transformation results
    \item Comparisons between the results generated by our framework and deep shapely portraits \cite{Xiao_2020_MM}
\end{itemize}

\subsection{Initialization Strategies}
\label{sec:init}
We provide additional analysis and examples to compare the reconstruction quality of embedded images using \textit{ResNet50} (used in the main paper) and mean-face (MF) initialization strategies. Quantitatively, the \textit{ResNet50} strategy produces slightly higher LPIPS scores than MF for images from both datasets according to Table~\ref{tbl:init-resnet-mean}. However, PSNR and SSIM scores of MF are equal or slightly higher than \textit{ResNet50} in both datasets. Given that LPIPS is more consistent with human perceptions than the other metrics \cite{zhang2018perceptual}, We chose \textit{ResNet50} as the initialization strategy in the latent space embedding step.

\begin{figure}[ht]
\centering
\includegraphics[width=0.9\linewidth]{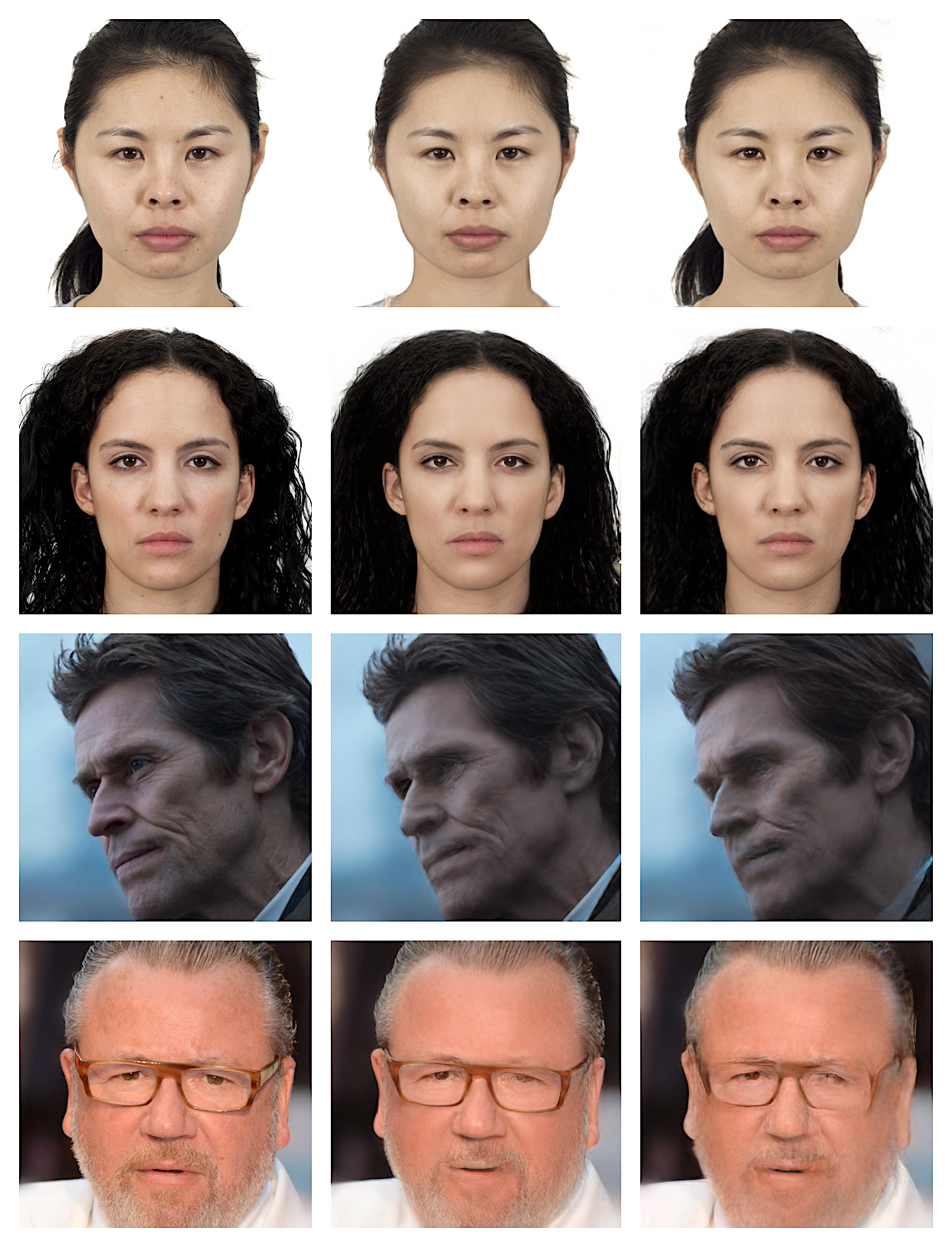}
\caption{Comparisons between \textit{ResNet50} and mean-face initialization strategies. Columns 1-3 show the original, \textit{ResNet50}-embedded, and mean-face-embedded images, respectively. Rows 1-2 and 3-4 display images from CFD-100 and WIDER-100 datasets, respectively. Zoom in for better resolution.}
\label{fig:init-resnet-mean}
\end{figure}

There are some cases where one strategy was able to generate marginally better results than the other and vice versa. For examples, in Fig.~\ref{fig:init-resnet-mean}, MF is more accurate in reconstructing a ponytail (row 1) and lip shape (row 2) of CFD-100 images than \textit{ResNet50}, whereas \textit{ResNet50} is better at reconstructing some edge cases (rows 3-4) in WIDER-100 than MF.

\begin{table}[ht]
\centering
\caption{Reconstruction quality scores for ResNet50 and Mean-Face Initialization Strategies}
\resizebox{0.8\linewidth}{!}{
\begin{tabular}{lccc}
\hline
\multicolumn{4}{c}{CFD-100} \\ \hline
 & \multicolumn{1}{l}{PSNR (dB) ($\uparrow$)} & \multicolumn{1}{l}{SSIM ($\uparrow$)} & \multicolumn{1}{l}{LPIPS ($\downarrow$)} \\ \hline
ResNet50 & 32.988 & 0.764 & 0.213 \\
Mean-Face & 33.023 & 0.768 & 0.216 \\ \hline
\multicolumn{4}{c}{WIDER-100} \\ \hline
ResNet50 & 31.625 & 0.747 & 0.312 \\
Mean-Face & 31.580 & 0.747 & 0.315 \\ \hline
\end{tabular}
}
\label{tbl:init-resnet-mean}
\end{table}

\subsection{Feature Entanglements}
\label{sec:entanglement}
We provide detailed description on the feature entanglement analysis from which the projected facial-weight vector was derived. In the main paper, we manually annotated 9.9K images (StyleGAN-9.9K) with facial-weight labels in order to extract the facial-weight direction in latent space via supervised learning. To investigate the feature entanglement problem, we measure correlations between the facial-weight direction and other facial attribute features. 

To achieve that, we followed similar procedures used in extracting the facial-weight vector. We first trained a binary classifier to predict \textit{age}, \textit{gender}, and \textit{mouth-opening expression} labels (one for each attribute). We selected relevant labeled face images from CelebA dataset \cite{liu2015faceattributes}, resulting in 3 sets of training data; each contains roughly 200K labeled images. For each set, we created a 0.875/0.125 train/test split and trained an attribute classifier by fine-tuning MobileNet (pre-trained on ImageNet). The accuracy scores of age, gender, and mouth-opening classifiers are 0.8863, 0.9353, and 0.8196, respectively. Next, we used the trained classifiers to assign the corresponding labels to StyleGAN-9.9K images. Then, we extracted the three attribute vectors using logistic regression trained on labeled StyleGAN-9.9K images (with 0.7/0.3 train-test split). The accuracy scores of the logistic regression classifiers for facial-weight, age, gender, and mouth-opening expression directions are 0.7993, 0.8034, 0.8312, and 0.7532, respectively. 

\begin{table}[ht]
\centering
\caption{Correlations between attribute directions}
\resizebox{0.95\linewidth}{!}{
\begin{tabular}{ccccc}
\hline
 & Facial weight & Gender & Age & Mouth open \\ \hline
Facial weight & 1.000 & -0.015 & -0.028 & 0.157 \\
Gender & - & 1.000 & -0.005 & -0.060 \\
Age & - & - & 1.000 & -0.117 \\
Mouth open & - & - & - & 1.000 \\ \hline
\end{tabular}
}
\label{tbl:attribute-correlation}
\end{table}

\begin{figure}[ht]
\centering
\includegraphics[width=0.9\linewidth]{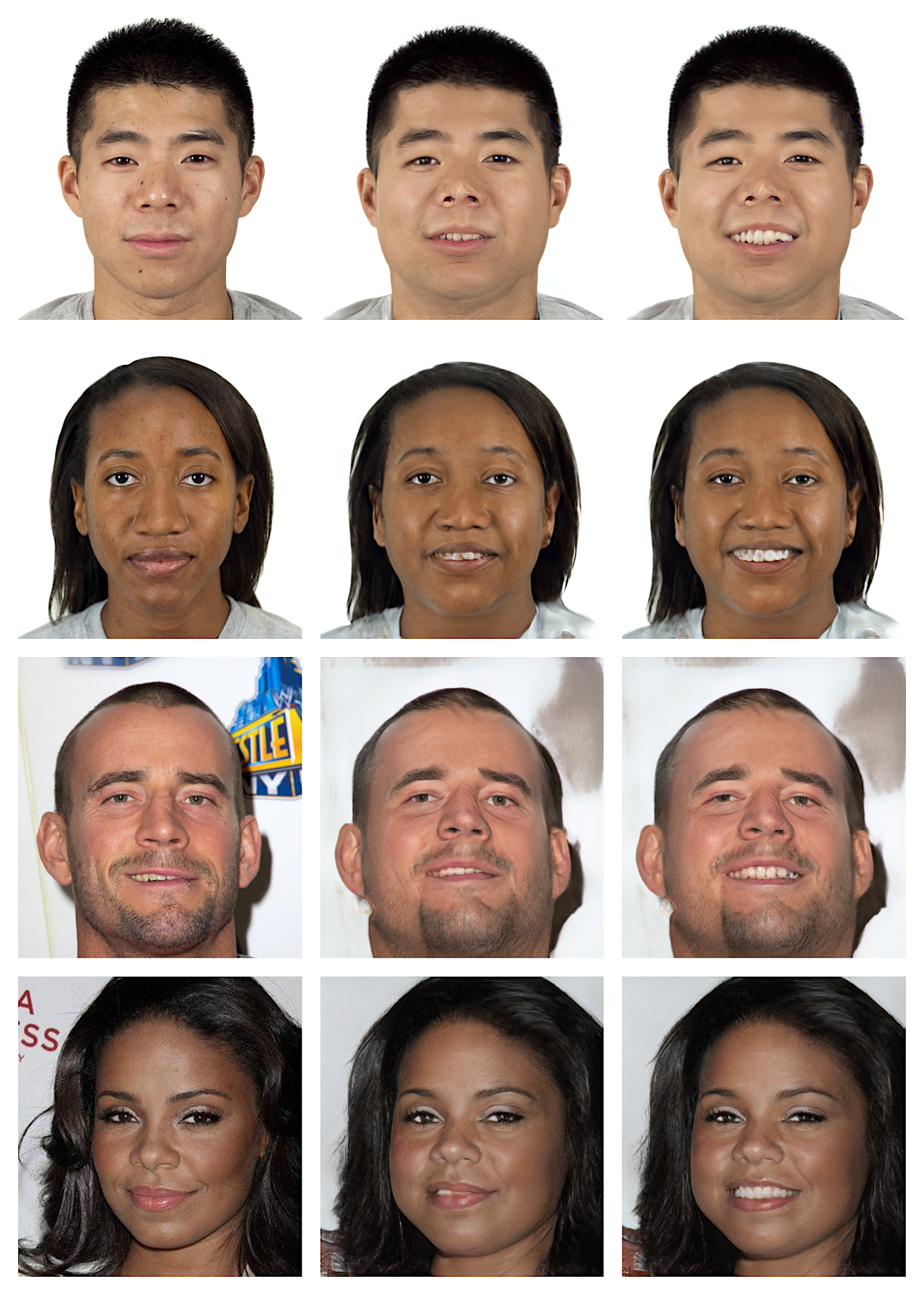}
\caption{Feature entanglement with (column 2) and without projected direction (column 3) at $\alpha=5$ given the input images in column 1}
\label{fig:mouth-comparison}
\end{figure}

Finally, we measured cosine similarity between the attribute vectors. As we can see in Table~\ref{tbl:attribute-correlation}, mouth-opening expression is more correlated with facial weight than the other attributes. As shown in Fig.~\ref{fig:mouth-comparison}, the subjects' mouths are more opened when the facial-weight attribute direction was not disentangled with the mouth-opening expression direction (column 3), compared to when projection subtraction was performed (column 2).

\subsection{Additional Transformation Results}
\label{sec:add_results}
\begin{figure*}[ht]
\centering
\includegraphics[width=\textwidth]{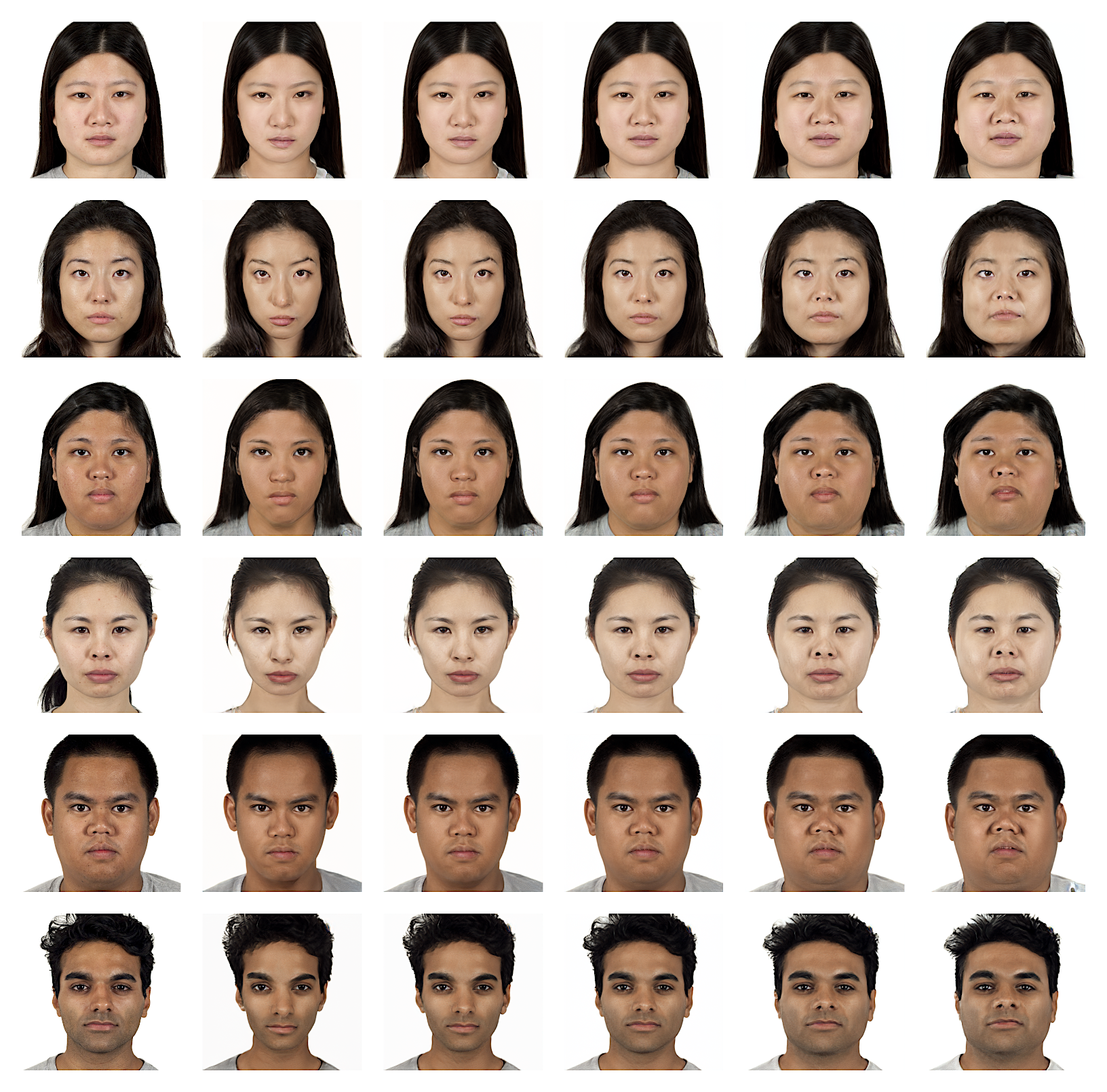}
\caption{Additional transformation results for Asian subjects in CFD-100. Columns 1 and 4 show the original and the embedded images, respectively. Columns 2-3 and 5-6 display the thinner ($\alpha=-5$ and $\alpha=-3$) and heavier ($\alpha=3$ and $\alpha=5$) transformations, respectively.}
\label{fig:cfd-01_results}
\end{figure*}

\begin{figure*}[ht]
\centering
\includegraphics[width=\textwidth]{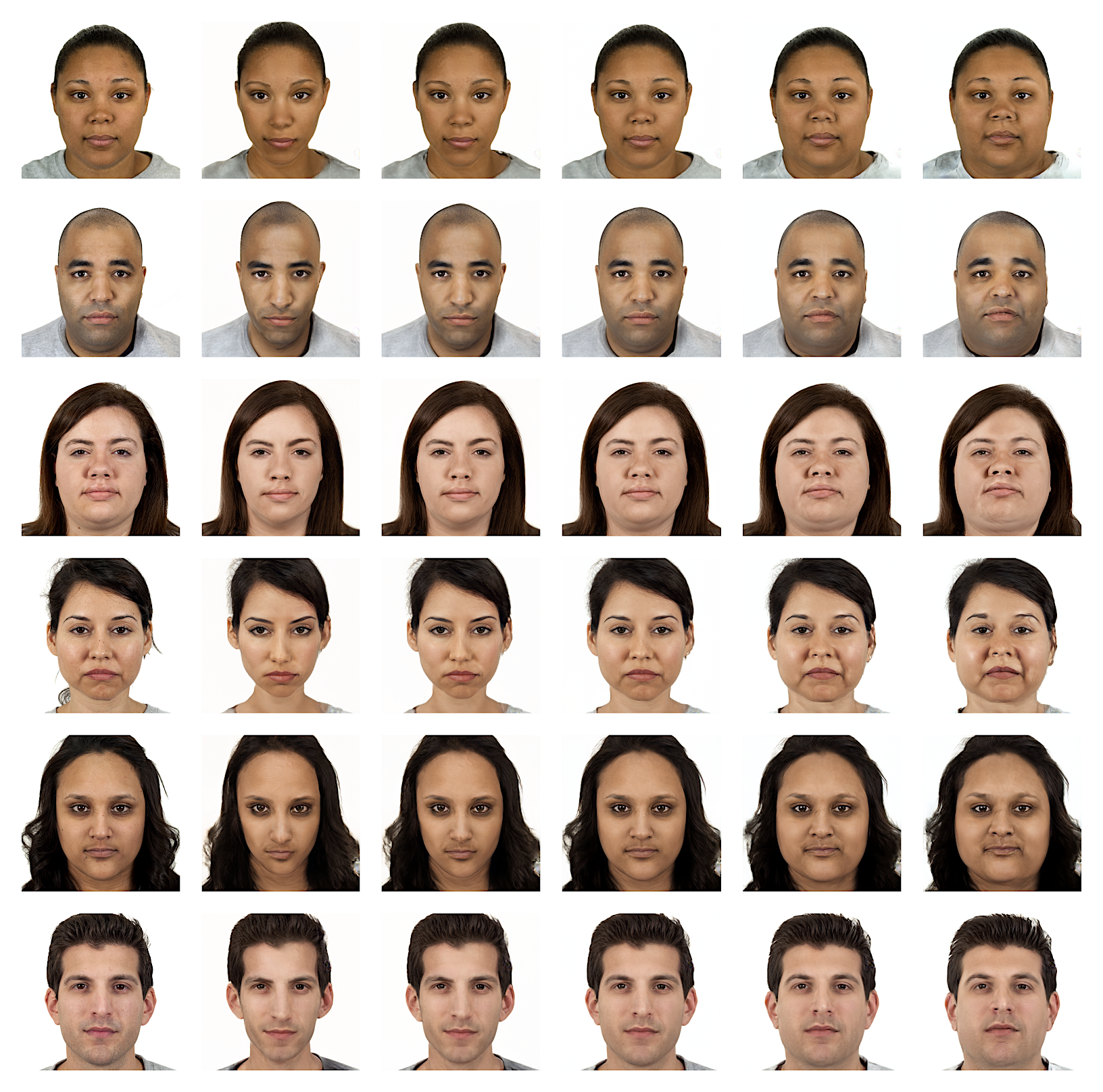}
\caption{Additional transformation results for Black and Latino subjects in CFD-100. Columns 1 and 4 show the original and the embedded images, respectively. Columns 2-3 and 5-6 display the thinner ($\alpha=-5$ and $\alpha=-3$) and heavier ($\alpha=3$ and $\alpha=5$) transformations, respectively.}
\label{fig:cfd-02_results}
\end{figure*}

\begin{figure*}[ht]
\centering
\includegraphics[width=\textwidth]{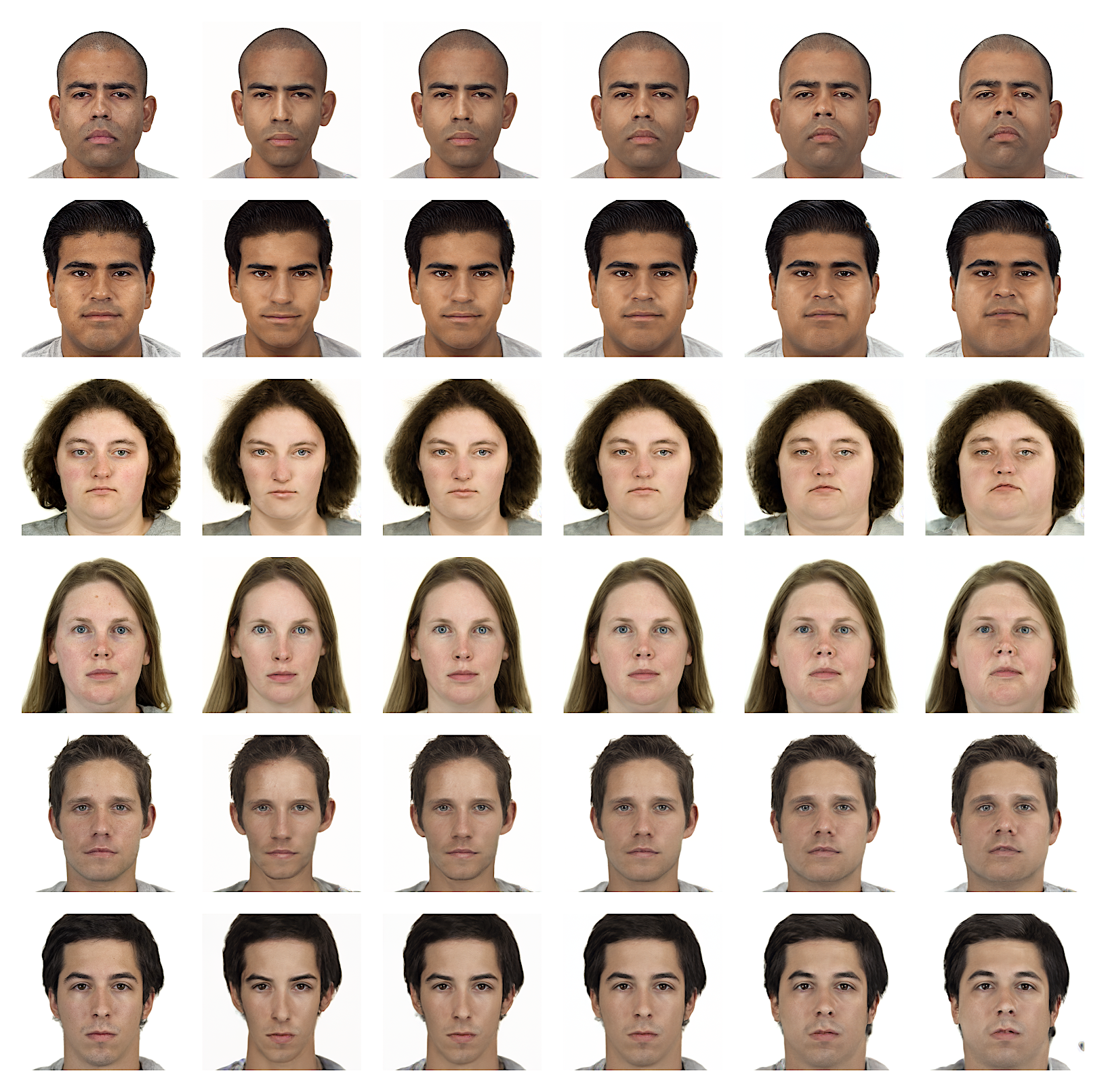}
\caption{Additional transformation results for Latino and White subjects in CFD-100. Columns 1 and 4 show the original and the embedded images, respectively. Columns 2-3 and 5-6 display the thinner ($\alpha=-5$ and $\alpha=-3$) and heavier ($\alpha=3$ and $\alpha=5$) transformations, respectively.}
\label{fig:cfd-03_results}
\end{figure*}

\begin{figure*}[ht]
\centering
\includegraphics[width=\textwidth]{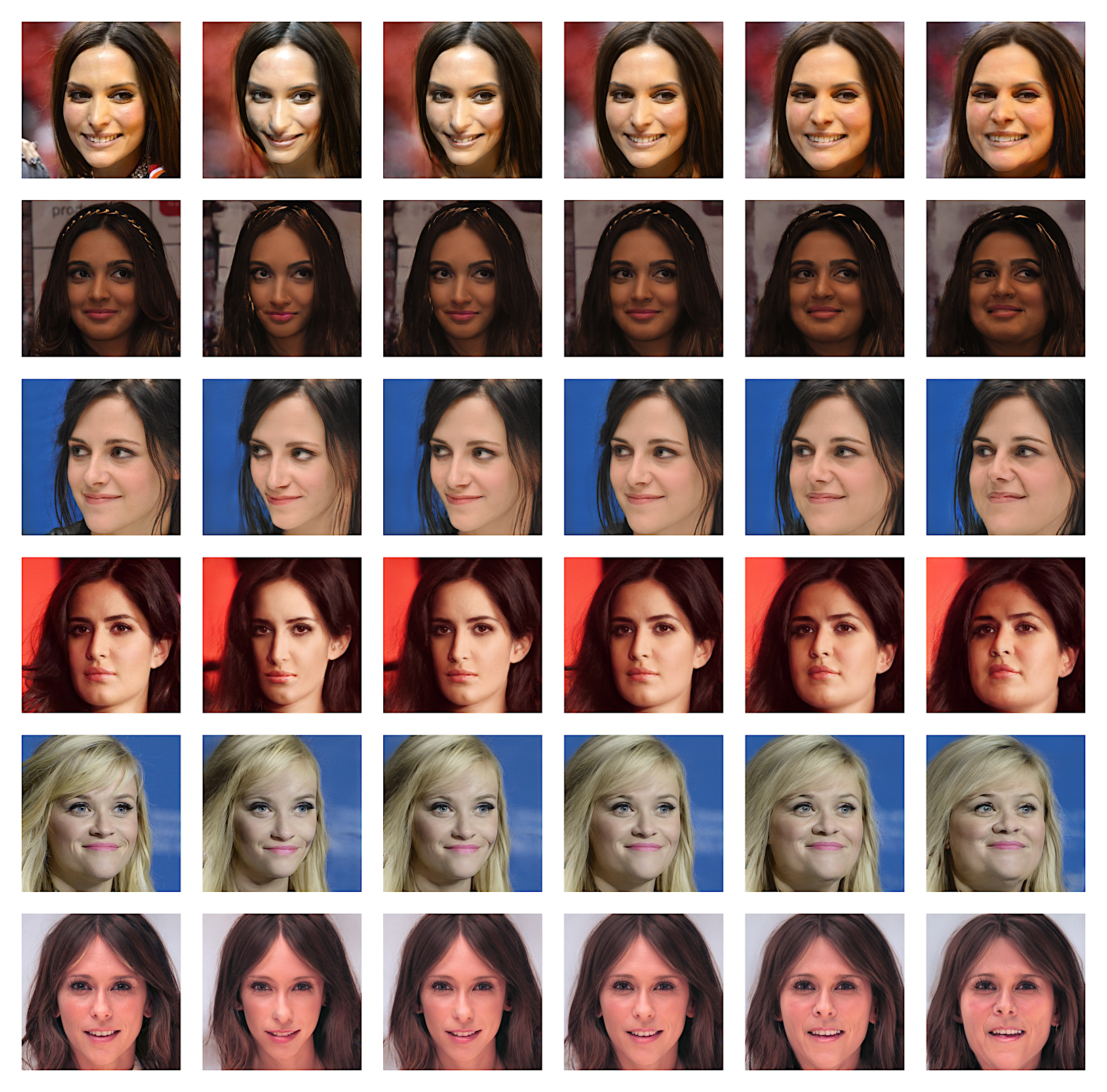}
\caption{Additional transformation results for subjects in WIDER-100. Columns 1 and 4 show the original and the embedded images, respectively. Columns 2-3 and 5-6 display the thinner ($\alpha=-5$ and $\alpha=-3$) and heavier ($\alpha=3$ and $\alpha=5$) transformations, respectively.}
\label{fig:wider-01_results}
\end{figure*}

\begin{figure*}[ht]
\centering
\includegraphics[width=\textwidth]{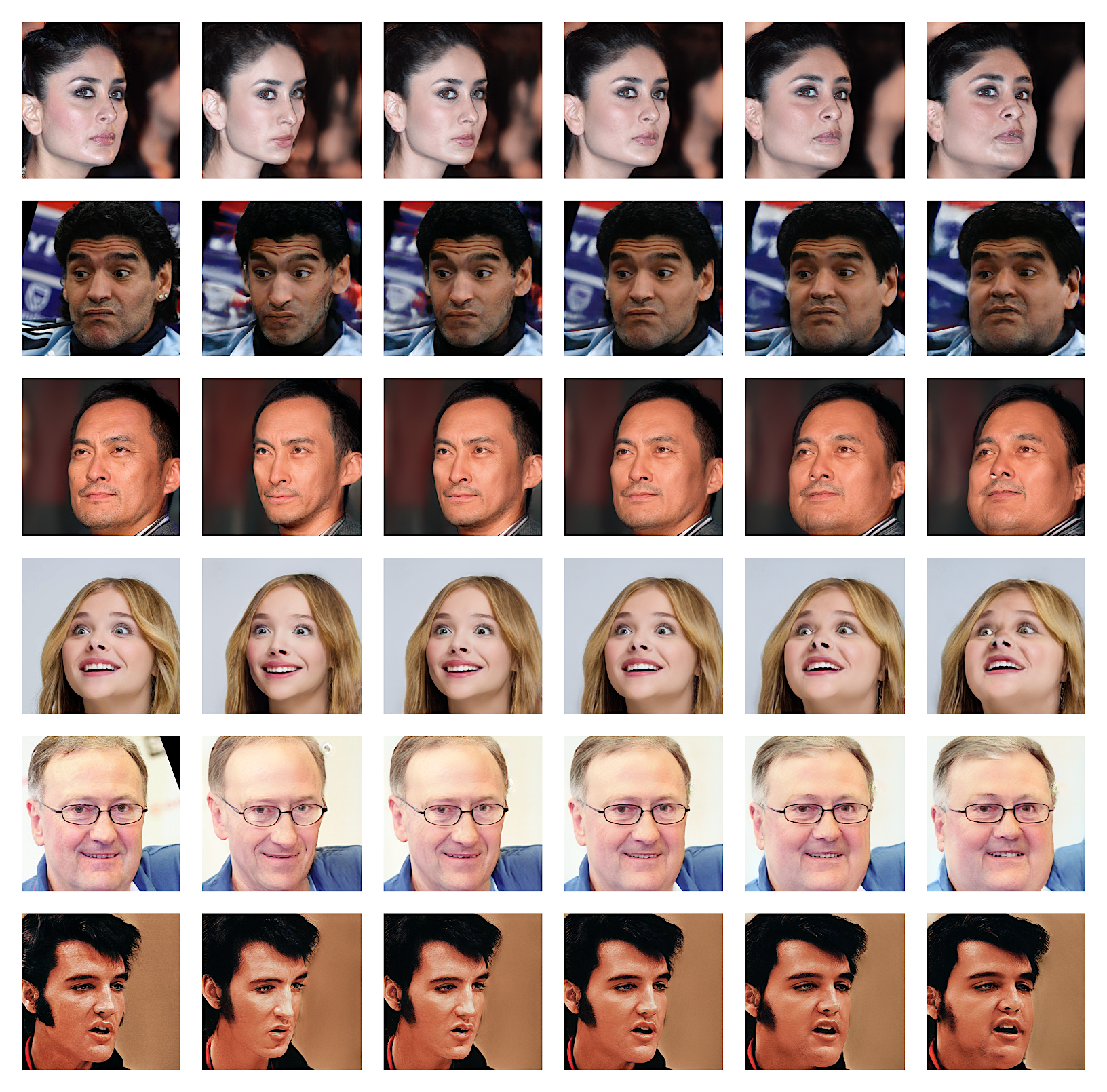}
\caption{Additional transformation results for subjects in WIDER-100. Columns 1 and 4 show the original and the embedded images, respectively. Columns 2-3 and 5-6 display the thinner ($\alpha=-5$ and $\alpha=-3$) and heavier ($\alpha=3$ and $\alpha=5$) transformations, respectively.}
\label{fig:wider-02_results}
\end{figure*}

\begin{figure*}[ht]
\centering
\includegraphics[width=\textwidth]{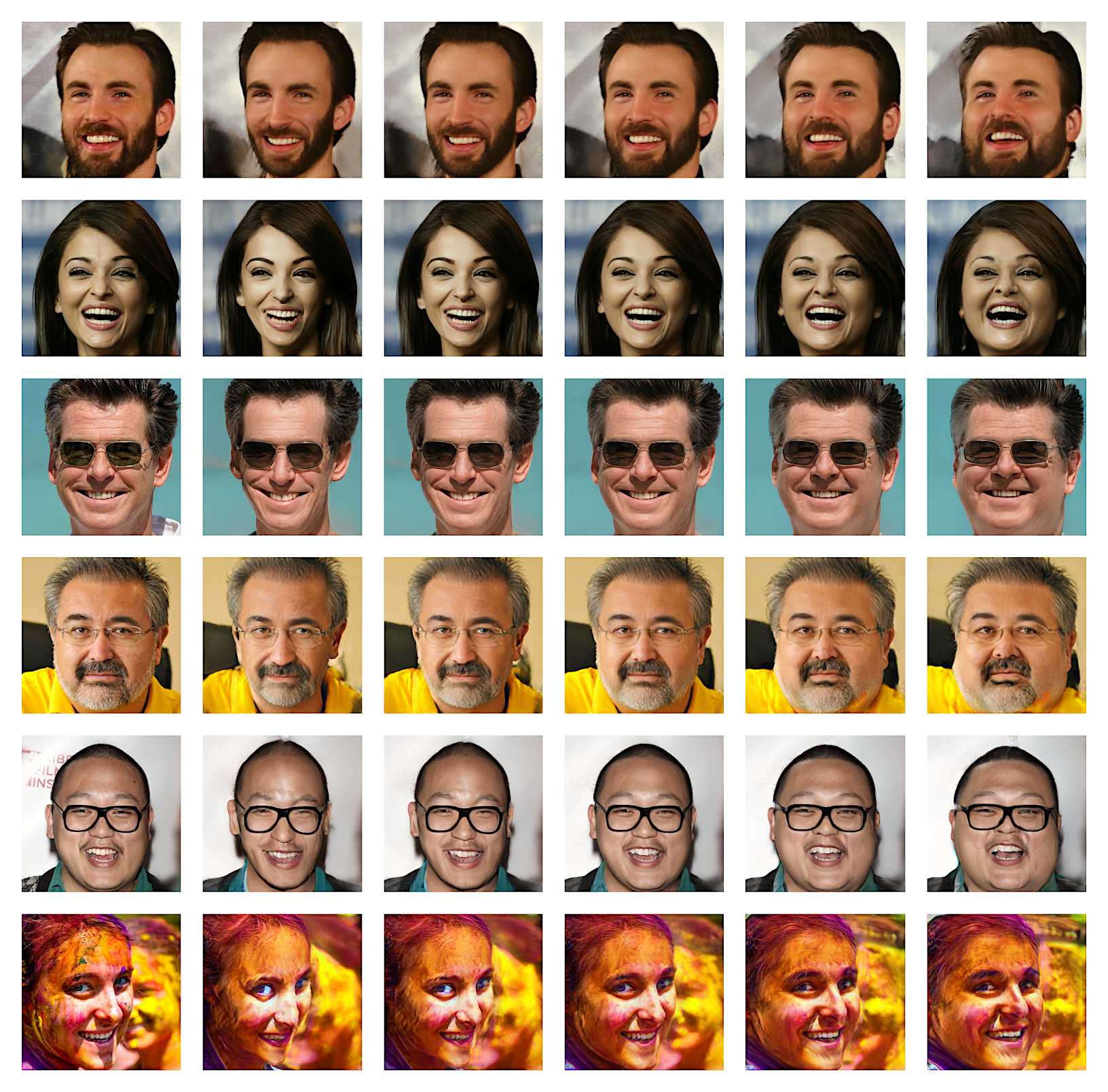}
\caption{Additional transformation results for subjects in WIDER-100. Columns 1 and 4 show the original and the embedded images, respectively. Columns 2-3 and 5-6 display the thinner ($\alpha=-5$ and $\alpha=-3$) and heavier ($\alpha=3$ and $\alpha=5$) transformations, respectively.}
\label{fig:wider-03_results}
\end{figure*}

We present additional examples of facial-weight transformation results supplementary to the results in the main paper. Fig.~\ref{fig:cfd-01_results}-\ref{fig:cfd-03_results} and \ref{fig:wider-01_results}-\ref{fig:wider-03_results} display additional transformation results for subjects in CFD-100 and WIDER-100 datasets, respectively. CFD-100 examples aim to show more variations in facial features, face shapes, and body weights within the same ethnicity, whereas WIDER-100 examples illustrate various facial expressions, face angles, and occlusions. 

\begin{figure*}[ht]
\centering
\includegraphics[width=\textwidth]{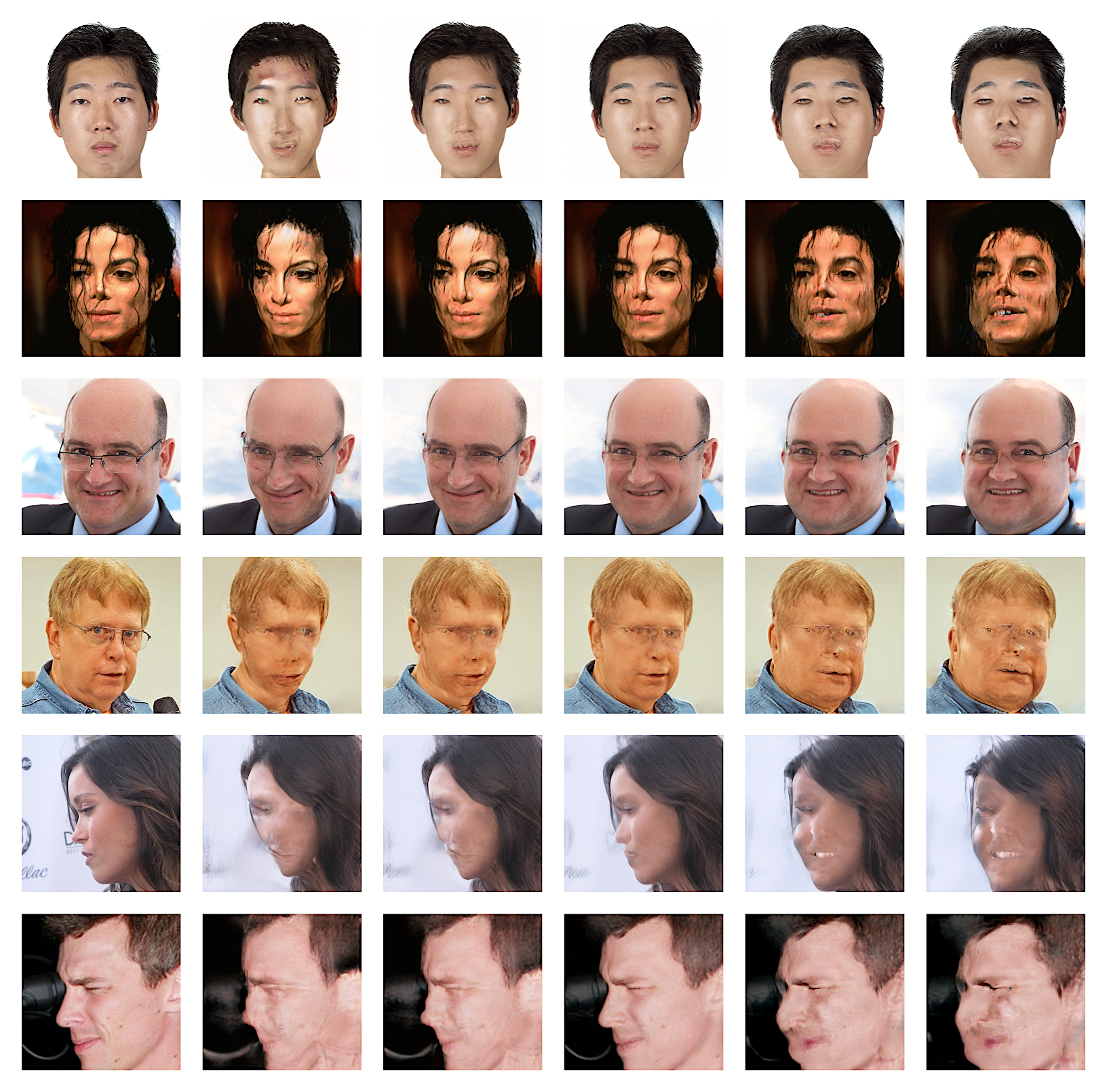}
\caption{Additional failure cases of face deformities. Columns 1 and 4 show the original and the embedded images, respectively. Columns 2-3 and 5-6 display the thinner ($\alpha=-5$ and $\alpha=-3$) and heavier ($\alpha=3$ and $\alpha=5$) transformations, respectively.}
\label{fig:fail-01_results}
\end{figure*}

\begin{figure*}[ht]
\centering
\includegraphics[width=\textwidth]{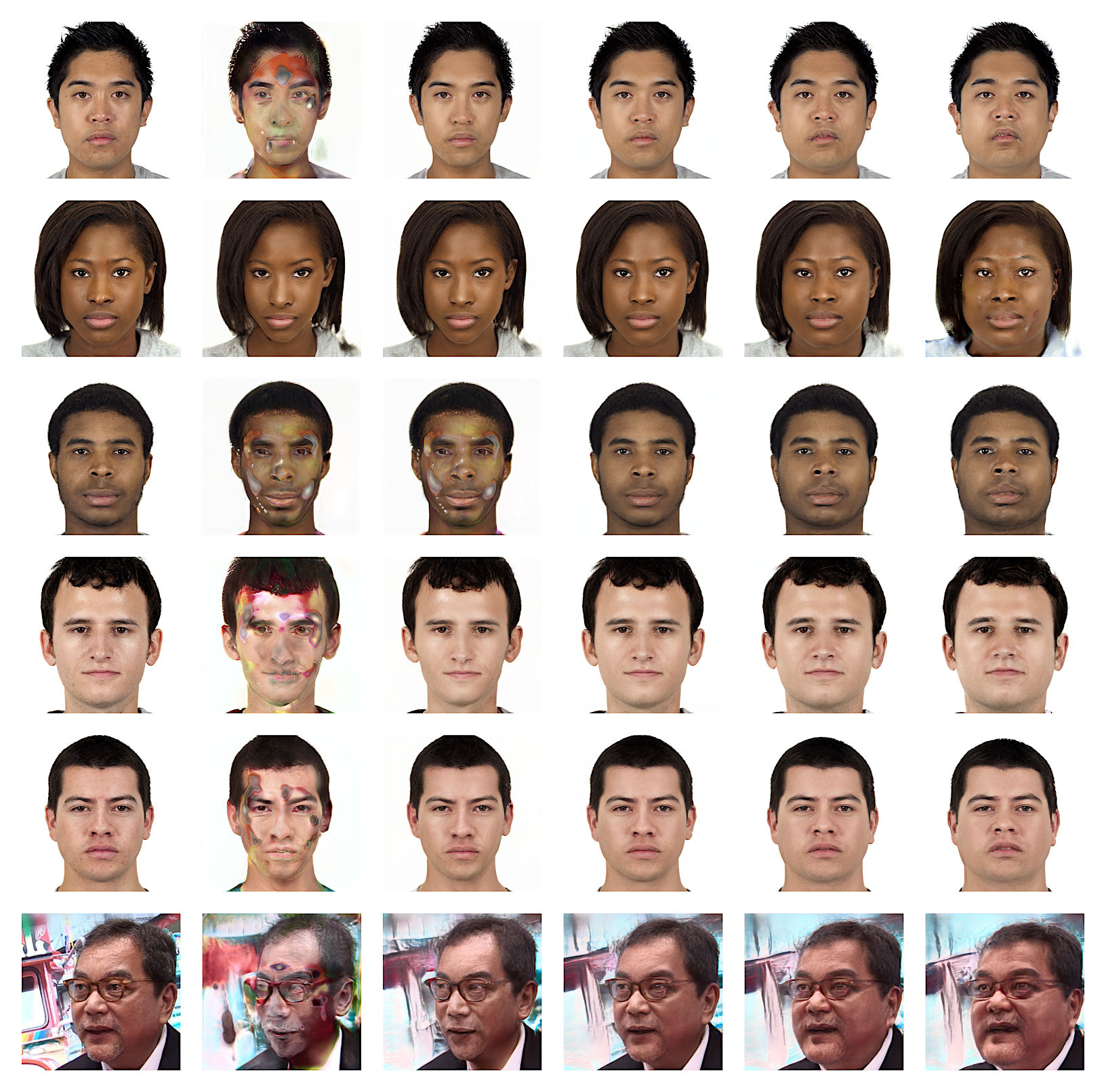}
\caption{Additional failure cases of blob artifacts. Columns 1 and 4 show the original and the embedded images, respectively. Columns 2-3 and 5-6 display the thinner ($\alpha=-5$ and $\alpha=-3$) and heavier ($\alpha=3$ and $\alpha=5$) transformations, respectively.}
\label{fig:blob-01_results}
\end{figure*}

\begin{figure*}[ht]
\centering
\includegraphics[width=\textwidth]{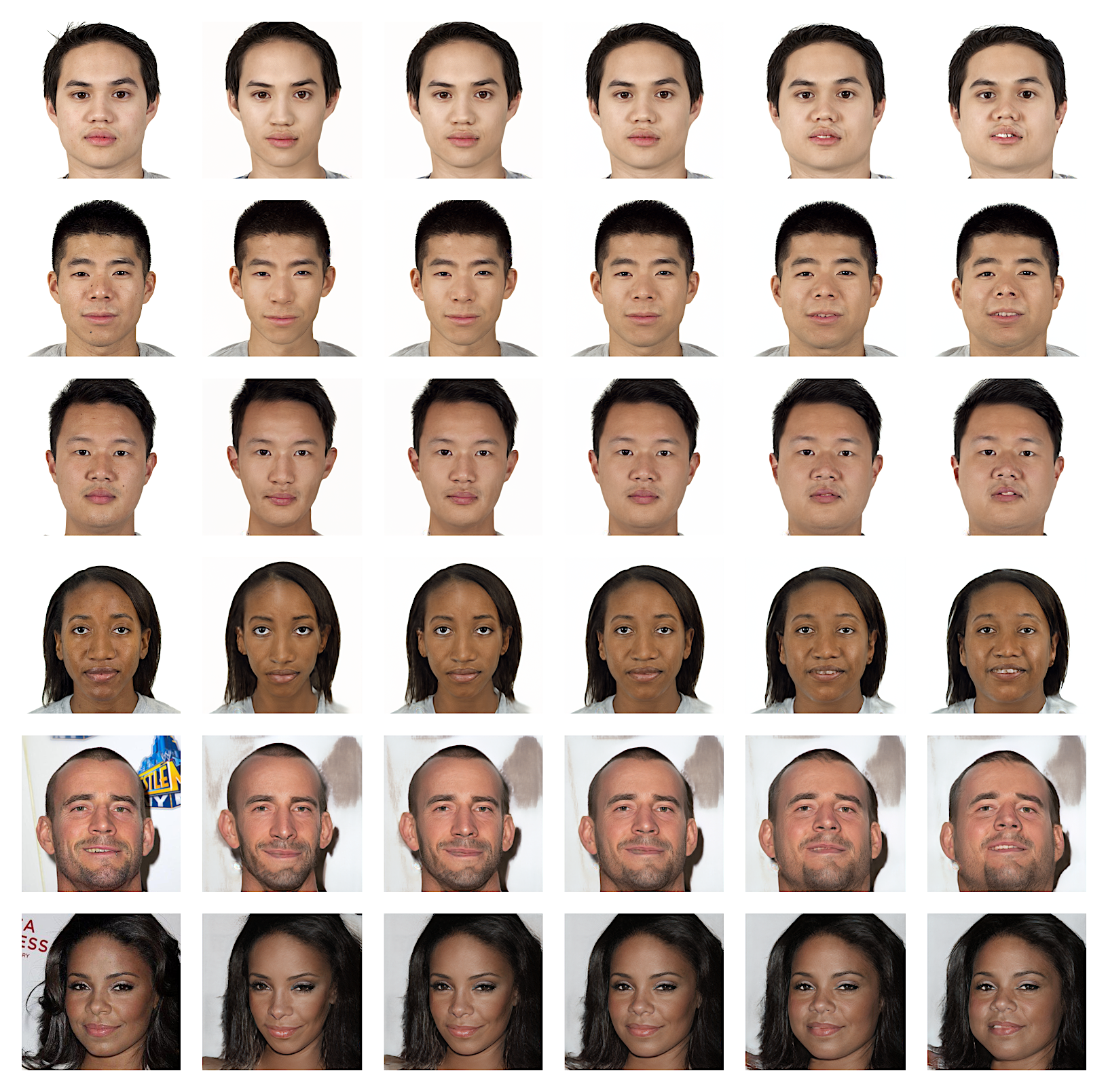}
\caption{Additional failure cases of mouth-opening feature entanglement. Columns 1 and 4 show the original and the embedded images, respectively. Columns 2-3 and 5-6 display the thinner ($\alpha=-5$ and $\alpha=-3$) and heavier ($\alpha=3$ and $\alpha=5$) transformations, respectively.}
\label{fig:mouth-01_results}
\end{figure*}

Fig.~\ref{fig:fail-01_results}-\ref{fig:mouth-01_results} show additional failure cases of face deformities, blob artifacts, and entanglement between facial weight and mouth-opening, respectively. Firstly, we can see in Fig.~\ref{fig:fail-01_results} that face deformities are caused by poor reconstruction quality (rows 1 and 4), occlusions such as hair covering face (row 2) and eyeglasses (row 3), and side-profile poses (rows 5-6). Secondly, Fig.~\ref{fig:blob-01_results} displays occurrences of blob artifacts in different transformation steps, especially when $\alpha<0$. Lastly, Fig.~\ref{fig:mouth-01_results} illustrate variations of mouth-opening expressions that correlate with heavier transformations ($\alpha>0$). As we can see, navigating along a positive facial weight direction ($\alpha>0$) sometimes causes the subject's mouth to open slightly.

\subsection{Comparisons with Deep Shapely Portraits}
\label{sec:dsp}
\begin{figure*}[ht]
\centering
\includegraphics[width=\textwidth]{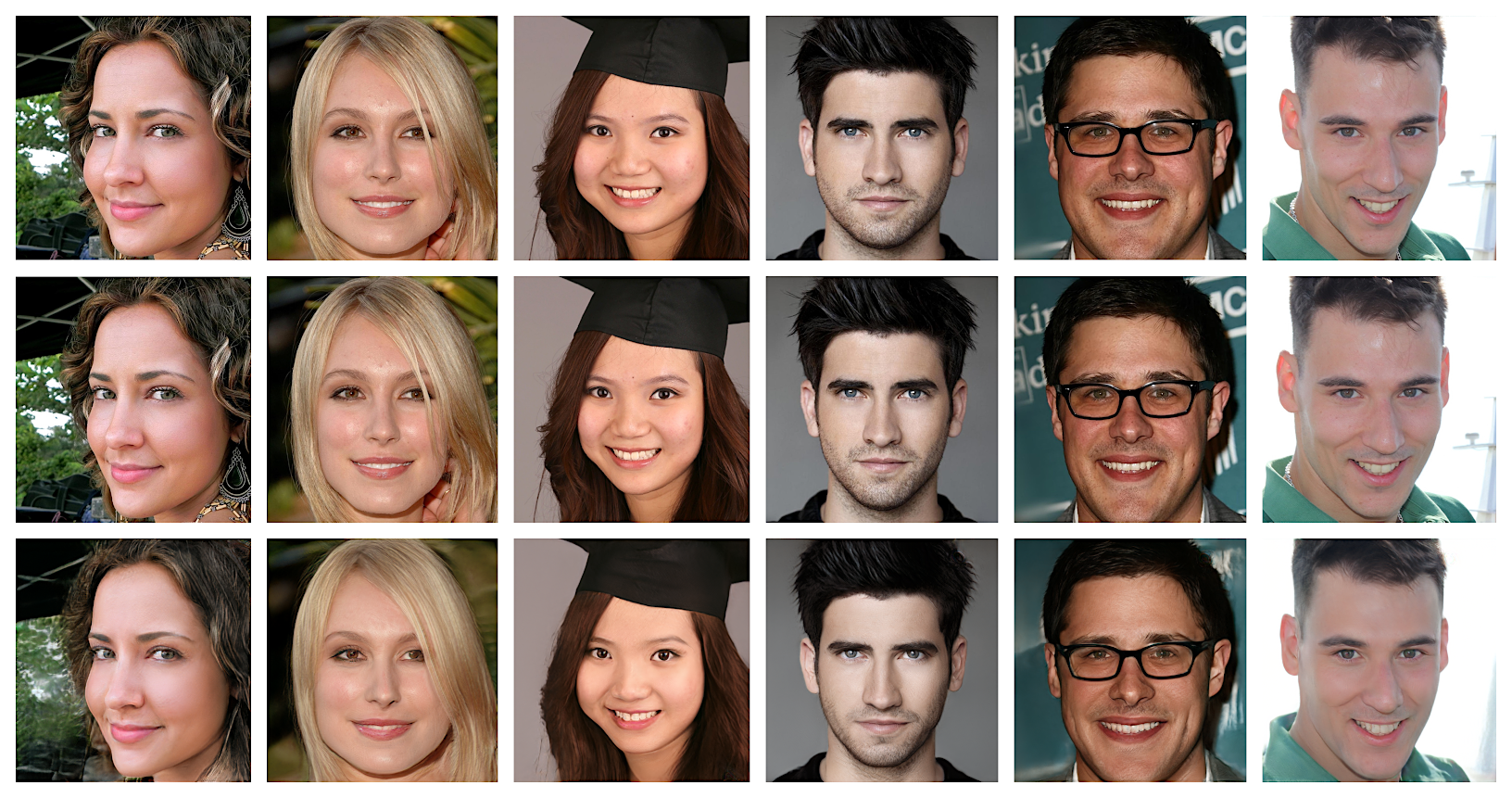}
\caption{Comparisons with deep shapely portraits. Columns 1 and 6 show six original images shown in Fig. 1 in Xiao et al.'s study \cite{Xiao_2020_MM}. Rows 1-3 display the original images (row 1), images generated by deep shapely portraits (row 2), and ours (row 3), respectively.}
\label{fig:dsp-01}
\end{figure*}

\begin{figure*}[ht]
\centering
\includegraphics[width=\textwidth]{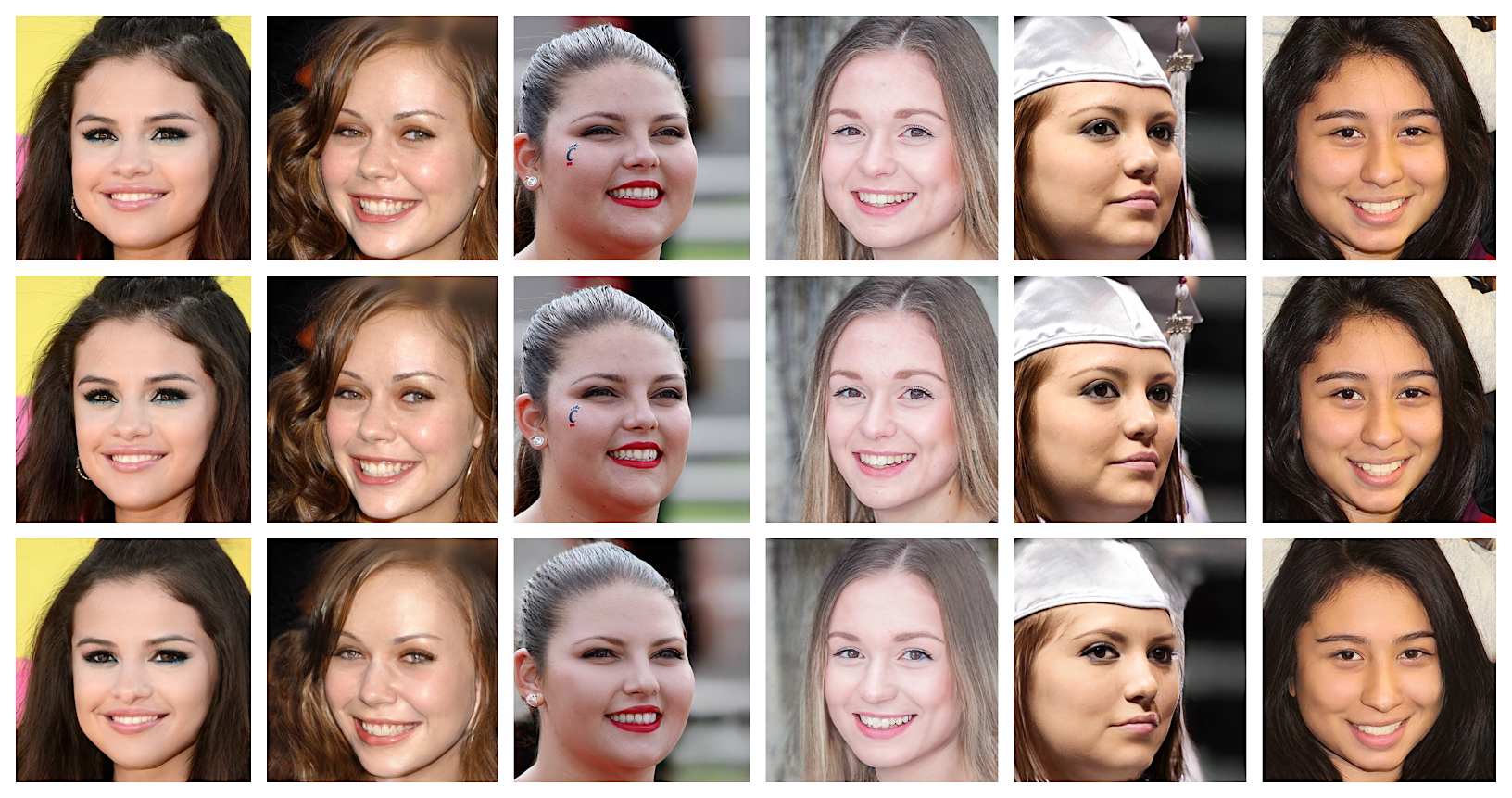}
\caption{Comparisons with deep shapely portraits. Columns 1 and 6 show six original images shown in Fig. 6 and 7 in Xiao et al.'s study \cite{Xiao_2020_MM}. Rows 1-3 display the original images (row 1), images generated by deep shapely portraits (row 2), and ours (row 3), respectively.}
\label{fig:dsp-02}
\end{figure*}

We provide additional results to qualitatively compare our framework with deep shapely portraits (DSP) \cite{Xiao_2020_MM}. DSP is a recent deep-learning based method utilizing sophisticated computer graphics techniques such as 3D face reconstruction, face reshaping, and warping to automatically transform face shapes of portrait images. As we can see in Fig.~\ref{fig:dsp-01} and \ref{fig:dsp-02}, the main advantage of DSP (row 2) over our framework (row 3) is in its ability to precisely extract and manipulate face shapes while preserving all other visual elements of the input images (row 1) as it does not recreate the whole image. Even though our framework was not able to accurately reconstruct accessories (e.g., earrings, face tattoo, tassel) and backgrounds without trading computation time for quality, the results demonstrate our framework's effectiveness in preserving the subjects' identity and facial expressions and generating face shapes closely resembling those of DSP, without relying on 3D models and explicit face reshaping functions.

\end{document}